\documentclass[runningheads]{llncs}
\usepackage[T1]{fontenc}

\usepackage{graphicx,verbatim}

\usepackage{subfig}

\usepackage{pifont}

\usepackage{amssymb}

\usepackage{amsmath}
\usepackage{tikz}
\newcommand{\xmark}{%
\tikz[scale=0.23] {
    \draw[line width=0.7,line cap=round] (0,0) to [bend left=6] (1,1);
    \draw[line width=0.7,line cap=round] (0.2,0.95) to [bend right=3] (0.8,0.05);
}}
\newcommand{\cmark}{%
\tikz[scale=0.23] {
    \draw[line width=0.7,line cap=round] (0.25,0) to [bend left=10] (1,1);
    \draw[line width=0.8,line cap=round] (0,0.35) to [bend right=1] (0.23,0);
}}

\usepackage{float}
\usepackage{graphicx}
\usepackage{caption}
\usepackage{array} 
\usepackage{booktabs}  

\usepackage{hyperref}
\usepackage{color}

\begin{document}
\title{ProtoEFNet: Dynamic Prototype Learning for Inherently Interpretable Ejection Fraction Estimation in Echocardiography}
\titlerunning{ProtoEFNet: Inherently Interpretable EF Estimation}

\author{Yeganeh Ghamary\inst{1}, Victoria Wu\inst{2}, Hooman Vaseli\inst{2}, Christina Luong\inst{3}, Teresa Tsang\inst{3}, Siavash A. Bigdeli\inst{1}, Purang Abolmaesumi\inst{2}}
\authorrunning{Y. Ghamary et al.}
\institute{Department of Applied Mathematics and Computer Science, Technical University of Denmark, Kongens Lyngby,
Denmark \and Department of Electrical and Computer Engineering, The University of British
Columbia, Vancouver, BC, Canada \and Vancouver General Hospital, Vancouver, BC, Canada \\
\email{\{s194258,sarbi\}@dtu.dk,\{victoriawu,hoomanv,purang\}@ece.ubc.ca}
\footnote{Y. Ghamary and V. Wu are joint first authors.\\ T. Tsang, S. A. Bigdeli and P. Abolmaesumi are joint senior authors. \\ Work done during Y. Ghamary’s external stay at UBC.}}

\maketitle
\begin{abstract}
Ejection fraction (EF) is a crucial metric for assessing cardiac function and diagnosing conditions such as heart failure. Traditionally, EF estimation requires manual tracing and domain expertise, making the process time-consuming and subject to inter-observer variability. Most current deep learning methods for EF prediction are black-box models with limited transparency, which reduces clinical trust. Some post-hoc explainability methods have been proposed to interpret the decision-making process after the prediction is made. However, these explanations do not guide the model’s internal reasoning and therefore offer limited reliability in clinical applications. To address this, we introduce ProtoEFNet, a novel video-based prototype-learning model for continuous EF regression. The model learns dynamic spatio-temporal prototypes that capture clinically meaningful cardiac motion patterns.
Additionally, the proposed Prototype Angular Separation (PAS) loss enforces discriminative representations across the continuous EF spectrum. Our experiments on the Echonet-Dynamic dataset show that ProtoEFNet can achieve accuracy on par with its non-interpretable counterpart while providing clinically relevant insight. The ablation study shows the proposed loss boosts the performance with a 2$\%$ increase in F1 score from $77.67\pm2.68$ to $79.64\pm2.10$. Our source code is available at: \url{https://github.com/DeepRCL/ProtoEF}.

\keywords{Ultrasound \and Echocardiography \and Ejection Fraction \and Regression \and Explainable AI \and Prototypical Neural Networks}

\end{abstract}

\section{Introduction}

Heart failure is a major global health issue that requires accurate and timely assessment for effective management. A key measure of cardiac function is ejection fraction (EF), which quantifies the percentage of blood pumped from the left ventricle with each heartbeat. EF plays a central role in diagnosing heart failure, guiding treatment, and predicting outcomes~\cite{huang2016accuracy,loehr2008heart}. It is typically measured using echocardiography, where clinicians manually trace the left ventricle at two key points in the cardiac cycle: end-diastole (ED) and end-systole (ES) to estimate volume changes~\cite{bamira2018imaging}. However, this process is highly operator-dependent, with inter-observer variation ranging from 7.6 to 13.9 percent, and requires significant expertise~\cite{beat_to_beat}. Automating ejection fraction estimation through artificial intelligence can enhance consistency, reduce clinician workload, and support large-scale screening.

Despite progress in automated EF estimation~\cite{kazemi2020bayes,lai2024echomen,maani2024coreecho,mokhtari2023gemtrans,mokhtari2022echognn,muhtaseb2022echocotr,beat_to_beat,reynaud2021shittybenchmark}, current methods face key challenges. Many rely on black-box deep learning models, offering limited transparency and reducing clinical trust. Explainability in medical AI is crucial, as clinicians require interpretable reasoning behind model predictions to facilitate adoption in medical settings. However, the existing explainable approaches use post-hoc techniques, such as attention weights~\cite{bahdanau2014neural,mokhtari2022echognn}, or gradients~\cite{selvaraju2017grad,smilkov2017smoothgrad}, that interpret the decision-making process after the prediction is made. These explanations are often inconsistent and do not inform the model’s internal reasoning, limiting their reliability in clinical applications.

To address these challenges, ante-hoc XAI methods have been introduced, embedding explainability directly into model architectures. Prototype-based models \cite{chen2019protopnet,kim2021xprotonet,kraft2021sparrow,vaseli2023protoasnet} are key examples, where networks learn class-specific prototypes that capture discriminative patterns explicitly used for classification. Unlike post-hoc methods that rely on abstract signals such as attention weights or gradients, which can be diffused, multi-layered, and difficult to interpret~\cite{chen2019protopnet,serrano2019attention}, prototypes offer more explicit, case-based explanations by grounding decisions in example-like visual features, mimicking a clinician's decision process. However, adapting prototype learning to ejection fraction estimation poses specific challenges. EF is a continuous value, requiring a regression-based approach, and unlike other assessments that use static images, echocardiographic assessment depends on capturing spatio-temporal information across frames.

Some prior work has extended prototype-based models to regression~\cite{hesse2024expert,hesse2022insightr} and video classification~\cite{vaseli2023protoasnet}. InsightRNet~\cite{hesse2022insightr} applies prototype learning to regression, but uses discrete ordinal labels rather than truly continuous targets. Furthermore, ~\cite{hesse2024expert,hesse2022insightr} rely on fixed-size patch-based prototypes, which are insufficient for capturing the complex and spatially diverse visual features of ejection fraction (EF), particularly the broad and varying motion of the LV wall.
\begin{figure}[t]
    \centering
    \includegraphics[width=\linewidth]{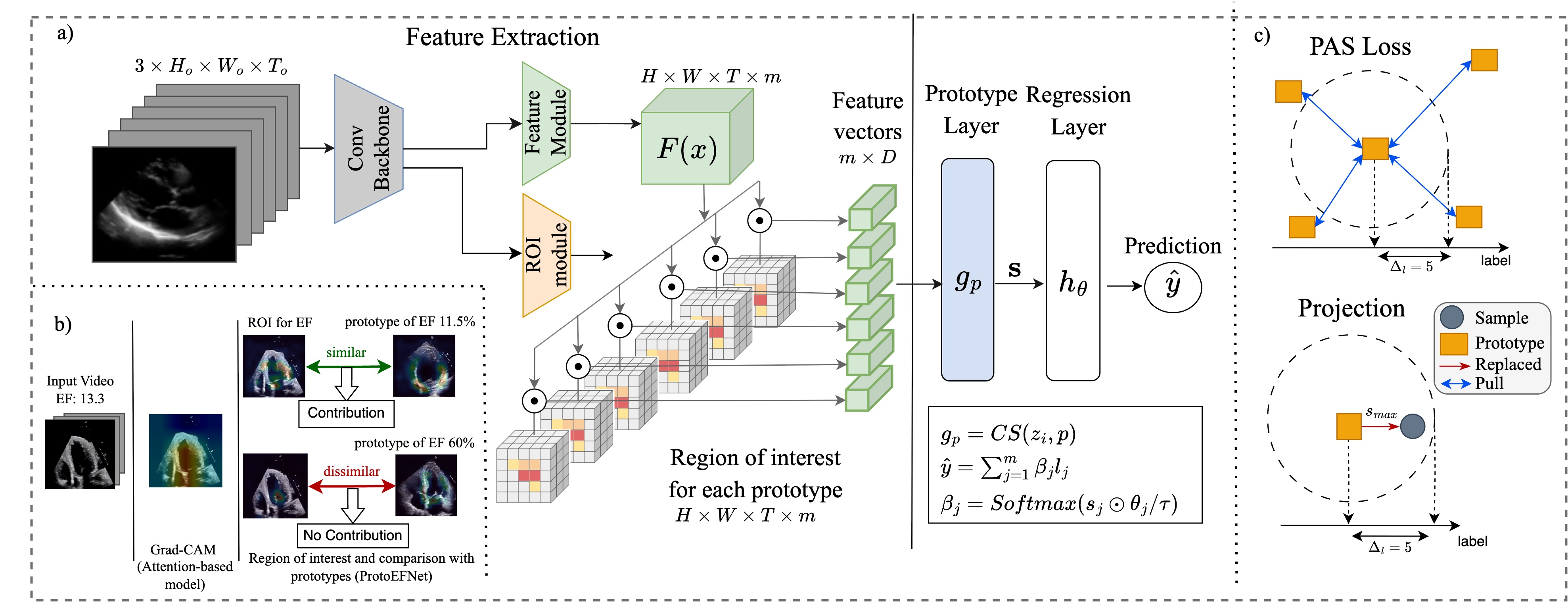}
    \caption{\textbf{(a)} An overview of the architecture of ProtoEFNet. The feature extractor uses an ROI module to focus on clinically relevant spatio-temporal regions, and the final prediction is the weighted sum of the prototype labels, \textbf{(b)} Grad-CAM attention of CoReEcho~\cite{maani2024coreecho}, and the decision process of ProtoEFNet. The activation maps on the input data and prototypes show where the model "looks at" when calculating the cosine similarity, \textbf{(c)} The Prototype Angular Separation (PAS) loss increases separation between prototypes with different EF ranges, and prototypes are projected to/replaced with the closest training sample within its EF range.}
    \label{fig1:methods:protoef}
\end{figure}
We introduce ProtoEFNet (Fig.~\ref{fig1:methods:protoef}), the first video-based prototype model for continuous regression and the first inherently interpretable approach to EF estimation. Our key contributions are: (1) learning dynamic spatio-temporal prototypes that capture clinically relevant motion patterns; (2) proposing a prototype angular separation (PAS) loss to enforce discriminative representations across the continuous EF spectrum; (3) achieving state-of-the-art explainability on the EchoNet-Dynamic dataset, and accuracy on par with leading black-box models; (4) demonstrating through visualizations that ProtoEFNet uniquely attends to essential cardiac features—such as LV wall motion and reduced mitral valve movement in EF$<40\%$ cases—whereas leading black-box methods produce diffuse or clinically irrelevant attention.

\section{Methods}

\subsubsection{Problem Definition}
We define a fixed number of $m$ learnable prototype vectors $\mathcal{P} = \{p_1, p_2, p_3, ..., p_m \}$, each associated with a continuous EF label $\mathbf{l} = \{l_1, ..., l_m\}$ and an \textit{importance score} $\mathbf{\theta} = \{\theta_1, ..., \theta_m\}$ that is learned during training. Based on the \textit{similarity scores} and the learned importance of each prototype, we generate a score sheet with scores indicating \textit{ prototype contributions}. Unlike in conventional classification tasks, where multiple prototypes are associated with each discrete class and scores are calculated at the class level~\cite{chen2019protopnet}, our approach assigns probabilities to individual prototypes, capturing the continuous nature of the label space.

\subsubsection{Feature Extractor} has a similar architecture to~\cite{vaseli2023protoasnet}, consisting of a pre-trained R(2+1)D-18 backbone, a feature module $F(.)$, and a Region of Interest (ROI) module $M_{p_k}(.)$. Given an input video $x \in \bbbr ^{H_{o} \times W_{o} \times T_{o} \times 3}$ with $T_{o}$ frames, the learned features are $F(x) \in \bbbr ^{H \times W \times T \times D}$, and the ROI module generates P occurrence maps $M_{p_k}(x) \in \bbbr ^{H \times W \times T}$ highlighting which regions of $F(x)$ are relevant when compared to $p_k$.  Occurrence maps highlight the regions in space and time where it is likely to observe relevant features for EF prediction. We perform a weighted average pooling of the spatiotemporal features using the learned occurrence maps as weights.

\subsubsection{Prototype Layer.} We use cosine similarity (CS) score as a similarity function between the features $f_{p_k}(x)$ and prototypes $p_k$, which are both D-dimensional vectors $s_k = g_{p_k}(f_{p_k}(x)) = CS(f_{p_k}(x), p_k)$.
\subsubsection{Regression Layer} is a linear layer with m weights denoted by \textbf{$\theta$}. Unlike class-based prototype models, the final prediction is calculated as a weighted average of the prototype labels:
\begin{equation}
    \hat{y} = \sum_{k=1}^{m} \beta_k \cdot l_k, \quad \beta_k(x) = \frac{e^{(s_k * \theta_k)/\tau}}{\sum_{i=1}^m e^{(s_k * \theta_k)/\tau}},
\end{equation}
where $\beta_k(.)$ indicates the \textit{contribution} of each prototype $p_k$ in the final prediction and is calculated using the softmax function of scaled similarity scores. To encourage \textit{sparse} explanations, we use a small temperature parameter of $\tau= 0.2$. This ensures that dissimilar prototypes have 0 contribution to the final prediction (see Figure~\ref{fig1:methods:protoef}(b)). The overview of ProtoEFNet can be seen in Figure~\ref{fig1:methods:protoef}(a).

\subsubsection{Training Algorithm.}The feature extractor, ROI module, regression layer, and prototype vectors are trained jointly. In the last epoch, the prototypes (and their labels) are \textit{projected} (replaced) onto the closest training feature (see Figure~\ref{fig1:methods:protoef}(c)). This allows us to visualize the prototypes, enhancing human-level explainability. We adapt the notion of class-based projection to the regression task using a threshold of $\Delta_l$. It is mathematically defined as: 
\begin{equation}
     p_j \leftarrow{} argmax_{z} CS(z, p_j), \qquad z=f_{p_j}(x_i) , |y_i - l_{j}| < \Delta_{l}.
     \label{eq:projection}
\end{equation} 

\subsubsection{Latent Space Regularization.} To learn representative prototypes, the latent space is learned using two distance-based losses. A regression-based cluster loss and prototype sample distance loss (PSD)~\cite{hesse2022insightr}. Cluster loss creates clusters around prototypes by pulling the samples with similar EF labels using a threshold of $\Delta_l$, and PSD loss ensures that there is at least one sample close to each prototype:
\begin{equation}
    \mathcal{L}_{Clst} = - \frac{1}{n} \sum_{i=1}^n \operatorname*{kmax}_{p \in \mathcal{P}^{c}} (s_{p}(i)), \quad  |y_i - l_j| < \Delta_{l},
\end{equation}
\begin{equation}
    \mathcal{L}_{PSD} = - \frac{1}{m} \sum_{j=1}^m log(1 - \min_{i \in [1, n]} \frac{d_{i,j}}{d_{max}}),
\end{equation}
where n is the batch size, $\operatorname*{kmax}$ is the maximum operation selecting the k closest prototypes to each sample, and $s_{p}(i)$ is the cosine similarity score between the prototype p and sample i. $\mathcal{P}^{c}$ visualizes the set of prototypes in the vicinity of the sample in the label space. $d_{max}$ is the maximum possible distance in the latent space, which is 2 for the cosine distance score, and $d_{i,j}$ is the cosine distance between sample i and prototype j.
Prototype models~\cite{chen2019protopnet,kim2021xprotonet} have been effective in classification by using separation losses to enforce clear boundaries between class-specific prototypes. However, in regression tasks, where outputs are continuous and ordered, preserving the ordinality and smooth transitions between clusters is essential. In the meantime, prototype vectors representing different EF regions should be further apart than those prototypes belonging to the same EF range. This ensures ordinality in the embedding space and that prototypes are semantically meaningful. However, in practice, we observed that even the most distant prototypes with EF labels of $15\%$ and $86\%$ have similar embedding vectors with a cosine similarity of around 0.7, suggesting that the embeddings lack meaningful distinction.
To address this, and inspired by~\cite{kraft2021sparrow}, we propose a novel prototype angular separation loss (PAS) that promotes greater inter-prototype distinction while respecting the continuity between the cluster of the data points required for regression tasks. The PAS loss addresses this by pulling prototypes from different EF regions further apart (see Figure~\ref{fig1:methods:protoef}(c). To measure the distance between prototypes, we utilize angular similarity—a monotonic transformation of cosine similarity that ranges from 0 to 1. The PAS loss can be defined as follows:
\begin{equation}
    \mathcal{L}_{PAS} = - \frac{1}{m} \sum_{i=1}^m \left[ \frac{1}{|\mathcal{P}^{\bar{c}}|} \sum_{j \in \mathcal{P}^{\bar{c}}} \log(1 - AS(p_{i}, p_{j})) \right], \quad \mathcal{P}^{\bar{c}} = \bigl\{ p_j \mid |l_i - l_j| > \Delta_{l} \bigr\},
\end{equation}
where $AS$ indicates angular similarity defined as:
\begin{equation}
    AS(\mathbf{p_i}, \mathbf{p_j}) = 1 - \frac{1}{\pi} \arccos(CS(\mathbf{p_i}, \mathbf{p_j})).
\end{equation}
$\mathcal{P}^{\bar{c}}$ is the set of prototypes that have EF label larger than $\Delta_{l}$ of the given prototype. Inside the bracket is the average of the log of the pairwise angular similarity. We chose the average due to its superiority over the max function.

\subsubsection{Occurrence Map Regularizor.} To further emphasize disease-specific areas, we incorporate the LV segmentation mask into the L1 regularization, penalizing activations outside the LV and reducing attention to irrelevant regions like the background. The overall cost function is defined below, where $\lambda$ represents weight of each loss term:  
\begin{equation}
    \mathcal{L} = \lambda_{MSE} \mathcal{L}_{MSE} + \lambda_{Clst} \mathcal{L}_{Clst} + \lambda_{PSD} \mathcal{L}_{PSD} +  \lambda_{PAS} \mathcal{L}_{PAS} + \lambda_{Occur} \mathcal{L}_{Occur}.
\end{equation}

\section{Experiment}
\subsubsection{Dataset and Implementation.} EchoNet-Dynamic~\cite{echonet} is the largest public echocardiogram dataset, containing 10,036 apical four-chamber (AP4) videos (112×112 resolution) with varying lengths, each labeled with a single EF value, LV segmentation, and end-systolic/diastolic frame indices. We follow the train-val-test split from~\cite{beat_to_beat} and frame sampling from~\cite{mokhtari2022echognn}. We augment the data using random rotation. Label balance was achieved by oversampling the minority region (EF $<50\%$), as addressing imbalance was beyond the scope of this study. The EF value outside the range of $[10\%,90\%]$ is clinically uncommon, and the dataset does not include any samples outside this range. Derived from the hyperparameter selection, each video is sampled as a clip of 64 frames (sampling period of 1) with the initial frame index sampled uniformly. 

All experiments were conducted using an NVIDIA A100 (40 GB) with PyTorch 2.0.1 and CUDA 12.8. Following hyperparameter tuning, we used 40 prototypes, a softmax temperature of $\tau = 0.2$ for regression layer, $\Delta_l = 5.0\%$, and $k = 3$ in $\mathcal{L}_{Clst}$. The prototype vectors were initialized randomly. Regression layer weights $\theta_h$ were initialized to 1, assigning equal importance to prototypes, and prototype labels were uniformly set from $10\%$ to $90\%$. This is to ensure that prototypes represent all regions of the label space, including low-density regions. We used Adam optimizer with a learning rate of $1e-4$ for the backbone and regression layer, $1e-3$ for feature and ROI modules and $3e-3$ for prototype vectors. We use pretrained weights of~\cite{kay2017kinetics} and fine-tuned jointly for 30 epochs with a batch size of 16. 
\subsubsection{Comparison with SOTA models.} In Table~\ref{tab:sota}, we compare ProtoEFNet with SOTA methods on EchoNet-Dynamic. We excluded~\cite{hesse2024expert,hesse2022insightr}, as adapting them to the dataset required major changes that altered their core architecture. We furthermore report the F1 score for the task of indicating whether EF values are lower than $40\%$, which is a strong indicator of heart failure~\cite{beat_to_beat}. ProtoEFNet outperforms most of the SOTA models, including the post-hoc explainable model~\cite{mokhtari2022echognn} and EchoCoTr with same frame size. The performance gap between ProtoEFNet and CoReEcho~\cite{maani2024coreecho} could potentially be narrowed with further training, and more extensive hyperparameter tuning. A key advantage of ProtoEFNet is its inherent explainability, offering transparent insights into its predictions.
\subsubsection{Explainability Analysis.} 
\begin{figure}[t]
    \centering
    \includegraphics[scale=0.27]{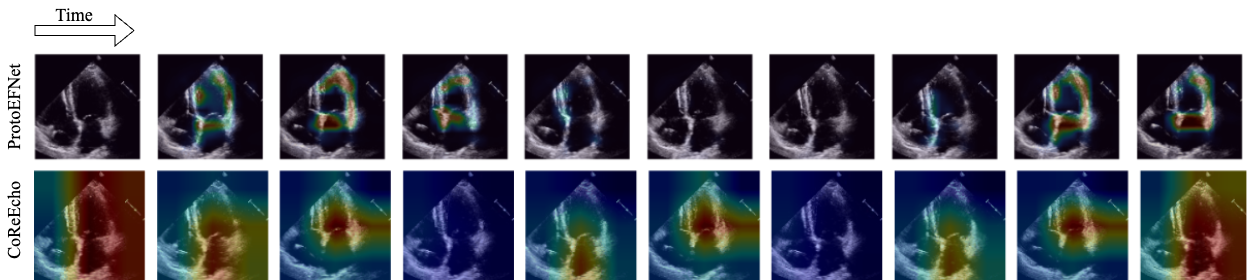}
    \caption{Grad-CAM~\cite{selvaraju2017grad} of CoReEcho (bottom row) and the activation map of ProtoEFNet (top row). ProtoEFNet is localised on LV wall motion and mitral valve movements during systole (contraction).}
    \label{fig:gradcam}
\end{figure}

\begin{figure}[t]
    \centering
    \includegraphics[scale=0.089]{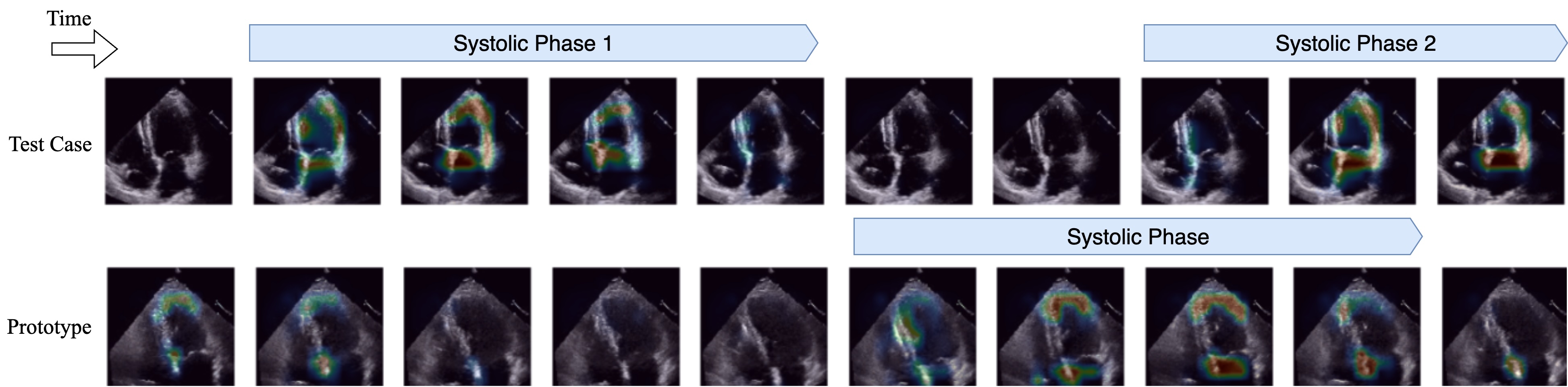}
    \caption{Activation maps of a test case and the top contributing prototype. The model captures periodic patterns and aligns spatio-temporal features. In this example, it assigns a high similarity score as it "looks at" the LV wall during systole in the input and identifies that it "looks like" the LV wall of the prototype during the same phase.}
    \label{fig:protomatch}
\end{figure}

Figure~\ref{fig1:methods:protoef}(b) illustrates ProtoEFNet’s decision process for a video sample, a more detailed description of the activation map visualisation can be found in \cite{kim2021xprotonet}. ProtoEFNet is \textit{inherently interpretable} and \textit{transparent}, with predictions directly linked to prototype contribution and similarity scores. Its explanations are \textit{sparse} and \textit{faithful}, relying only on prototypes with EF values close to the true label—prototypes with distant EF values (e.g., 60$\%$) do not influence the prediction for the sample with EF of 13$\%$. Activation maps on the input video demonstrate both spatial (anatomical localization) and temporal (motion) explainability, highlighting \textit{clinically relevant} features such as reduced LV wall motion and mitral valve movement during systole. Prototypical features shown as activation maps on top of prototypical cases show distinct EF-related patterns: the 60$\%$ EF prototype exhibits healthy LV and MV motion, while the 11$\%$ EF prototype shows reduced LV motion and a thin LV wall, both clinically meaningful. In Figure~\ref{fig:gradcam}, CoReEcho’s spatio-temporal attention (Grad-CAM~\cite{selvaraju2017grad}) is compared to ProtoEFNet’s activation map. ProtoEFNet demonstrates superior localization, focusing on key structures like the LV wall and mitral/aortic valves, while CoReEcho and EchoCoTr highlight non-specific or clinically irrelevant regions, such as background (see~\cite{maani2024coreecho}). ProtoEFNet also learns the periodic nature of echo videos and correctly aligns the spatio-temporal features of the input clip to those of the prototype, seen in Figure~\ref{fig:protomatch}.
\begin{table}[t]
\caption{The quantitative results on the EchoNet-Dynamic test set. \textbf{P} indicates the frame sampling period. Most approaches average clip-level predictions over the entire video (\textbf{EV}), while others use a single heartbeat (\textbf{SH}). Results marked with (*) denote reproduced performance, while ($^1$) indicates the SOTA method with the frame sampling similar to ours.}\label{tab:sota}
\begin{tabular}{l l l l l l l l l}
\hline
Method & Frames & P & Clips & R2 $\uparrow$ & MAE $\downarrow$ & RMSE $\downarrow$ & $F1_{<40\%}$$\uparrow$ & Explainable\\
\hline
Reynaud et al.~\cite{reynaud2021shittybenchmark} & 128 & 1 & SH & 52 &  5.59 & 8.38 & NA & \xmark\\
Bayesian~\cite{kazemi2020bayes}  & 32 & 1 & SH & 75 & 4.46 & NA & 77 & \xmark \\ 
EchoCoTr~\cite{muhtaseb2022echocotr}$^{1}$ & 36 & 2 & EV & 79 & 4.18 & 5.59 & NA & \xmark\\
EchoCoTr~\cite{muhtaseb2022echocotr} & 36 & 4 & EV & 81 & 3.98 & 5.34 & NA & \xmark\\
CoReEcho~\cite{maani2024coreecho}  & 36 & 1 & 3 Clips & \textbf{82} & \textbf{3.90} & \textbf{5.13} & \textbf{80} & \xmark\\
GEMTrans~\cite{mokhtari2023gemtrans}$^1$ & 64 & 1 & EV & 79 & 4.15 & NA & NA & \xmark\\
Resnet2+1D~\cite{beat_to_beat}$^1$* & 64 & 1 & EV & 80  & 4.10 & 5.47 & 78 & \xmark \\
\hline
EchoGNN~\cite{mokhtari2022echognn}$^1$ & 64 & 1 & EV & 76 & 4.45 & NA & \textbf{78} & post-hoc\\
ProtoEFNet (ours)& 64 & 1 & EV & \textbf{80} & \textbf{4.07} & \textbf{5.47} & \textbf{78} & \cmark \\
\hline
\end{tabular}
\end{table}
\begin{figure}[t]
    \centering
    \includegraphics[scale=0.6]{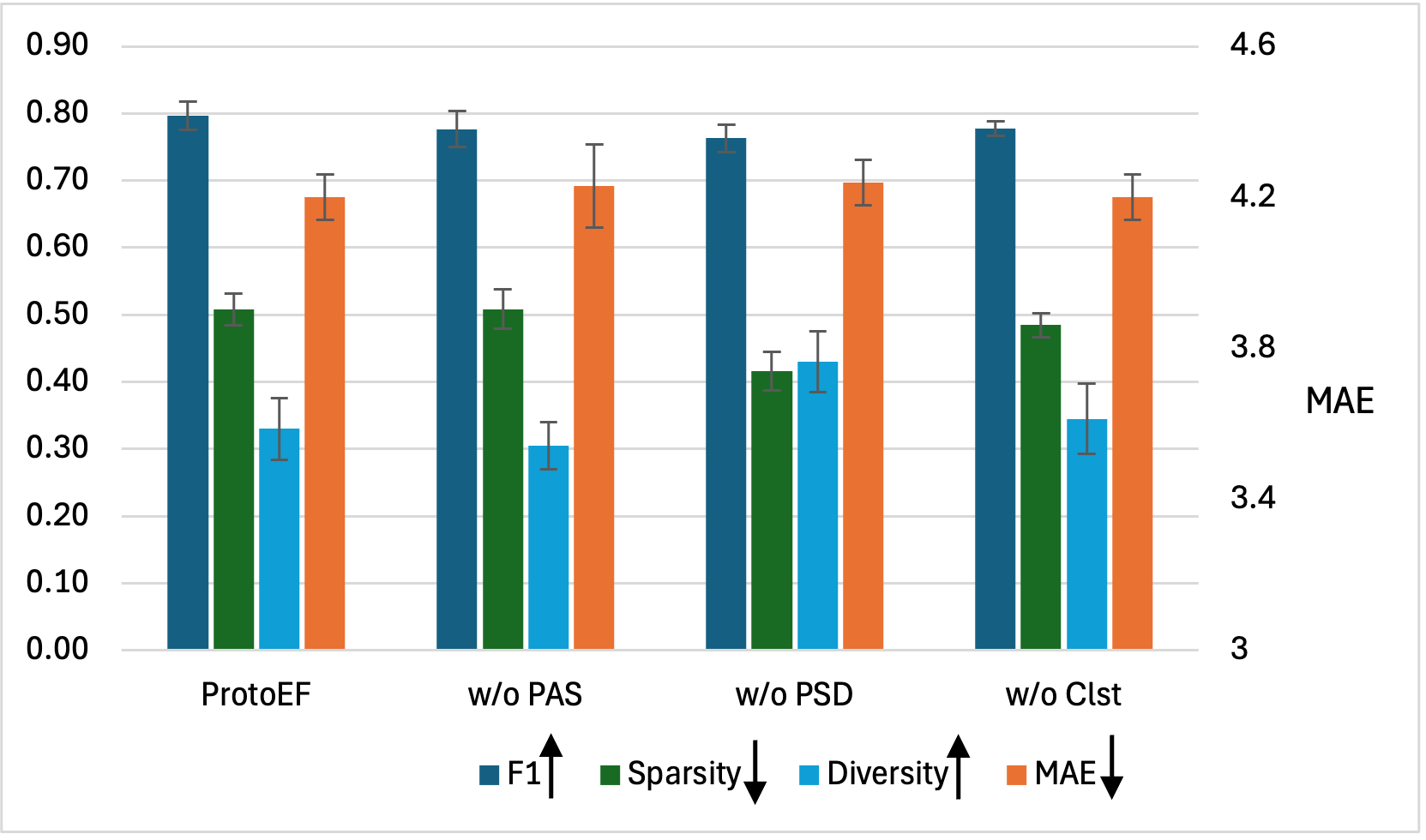}
    \caption{Ablation Study of different loss components on validation set. Standard deviation is calculated across 5 repetitions of each experiment.}
    \label{fig:ablation}
\end{figure}
\subsubsection{Ablation Study.} 
\begin{figure}[t]
    \centering
    \includegraphics[scale=0.05]{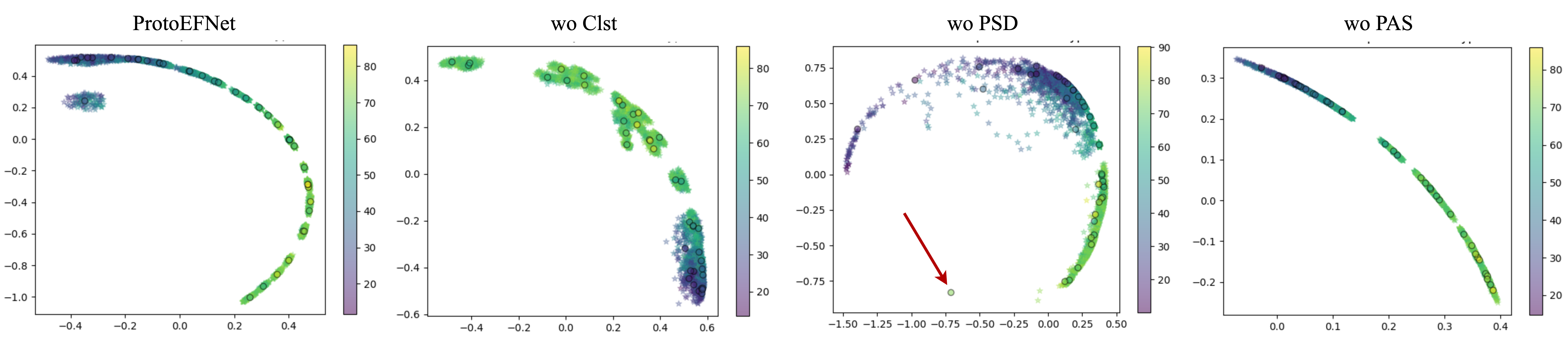}
    \caption{The PCA plots of the prototypes (circles) and the top 100 closest latent features of the validation set to each prototype (stars). The colors indicate the ground truth EF values.}
    \label{fig:loss_ablation:pca}
\end{figure}

Figure~\ref{fig:ablation} shows an ablation study of different loss components. Including all loss components yields the best regression performance based on MAE and F1 scores. We evaluate prototype quality using Sparsity and Diversity metrics~\cite{hesse2022insightr} scaled by the number of prototypes: effective explanations rely on the contribution of a few prototypes (\textit{low Sparsity}), but different predictions rely on different prototypes (\textit{high Diversity}). To demonstrate the effect of these components in the embedding space, we visualize 2D PCA plot of the prototypes and the 100 closest validation features in Figure \ref{fig:loss_ablation:pca}. Removing $\mathcal{L}_{PSD}$ improves diversity and sparsity but degrades regression performance, as reflected by F1 and MAE scores. Moreover, some prototypes become outliers with no nearby training samples (see Figure~\ref{fig:loss_ablation:pca}). Without $\mathcal{L}_{Clst}$, the samples and prototypes are scattered in the embedding space without any clear ordinality. $\mathcal{L}_{PAS}$ decreases MAE from $4.23\pm 0.11$ to $4.20\pm 0.06$, and increases F1 score from $77.67\pm 2.68$ to $79.64\pm 2.10$, while producing more distinct prototypes and ordinally structured clusters in the embedding space.
\section{Conclusion}
We proposed ProtoEFNet, the first prototype-based model for video-based continuous EF regression. ProtoEFNet has superior performance to the \textit{post-hoc} explainable model~\cite{mokhtari2022echognn} and a performance on par with the black-box models, while offering inherent interpretability, transparency, and clinically meaningful explanations. Qualitative analysis shows superior focus on key cardiac structures, and ablation results confirm the effectiveness of the proposed Prototype Angular Separation loss. Future work will address prototype learning for uncommon EF ranges (minority region).

\subsubsection{Acknowledgements.} This work was supported in part by the Technical University of Denmark's Travel Grant, Marie og Anders Manssons Memorial Grant, the Canadian Institutes of Health Research (CIHR), and the Natural Sciences and Engineering Research Council of Canada (NSERC). Resources were provided through DTU Computing Center at Technical University of Denmark~\cite{DTU_DCC_resource} and Advanced Research Computing at the University of British Columbia~\cite{UBC_SOCKEYE}.

\bibliographystyle{splncs04}
\bibliography{mybibliography}
\end{document}